\pdfoutput=1

\documentclass[11pt]{article}

\usepackage{ACL2023}

\usepackage{times}
\usepackage{latexsym}

\usepackage[T1]{fontenc}

\usepackage[utf8]{inputenc}

\usepackage{microtype}

\usepackage{inconsolata}

\usepackage{graphicx}
\usepackage{caption}
\usepackage{subcaption}
\usepackage{arydshln}
\usepackage{multirow}
\usepackage{enumitem}

\usepackage{pifont}
\newcommand{\cmark}{\ding{51}}%
\newcommand{\heart}{\ensuremath\heartsuit}

\title{Explaining Relationships Among Research Papers}

\author{Xiangci Li ~~ Jessica Ouyang\\
  Department of Computer Science \\
  University of Texas at Dallas \\
  Richardson, TX 75080 \\
  \tt lixiangci8@gmail.com, \\ 
  \tt Jessica.Ouyang@UTDallas.edu
}

\begin{document}
\maketitle
\begin{abstract}
Due to the rapid pace of research publications, keeping up to date with all the latest related papers is very time-consuming, even with daily feed tools. There is a need for automatically generated, short, customized literature reviews of sets of papers to help researchers decide what to read. While several works in the last decade have addressed the task of explaining a single research paper, usually in the context of another paper citing it, the relationship among multiple papers has been ignored; prior works have focused on generating a single citation sentence in isolation, without addressing the expository and transition sentences needed to connect multiple papers in a coherent story. In this work, we explore a feature-based, LLM-prompting approach to generate richer citation texts, as well as generating multiple citations at once to capture the complex relationships among research papers. We perform an expert evaluation to investigate the impact of our proposed features on the quality of the generated paragraphs and find a strong correlation between human preference and integrative writing style, suggesting that humans prefer high-level, abstract citations, with transition sentences between them to provide an overall story.
\end{abstract}

\begin{figure*}
     \centering
     \begin{subfigure}[b]{0.48\textwidth}
         \centering
         \includegraphics[width=\textwidth]{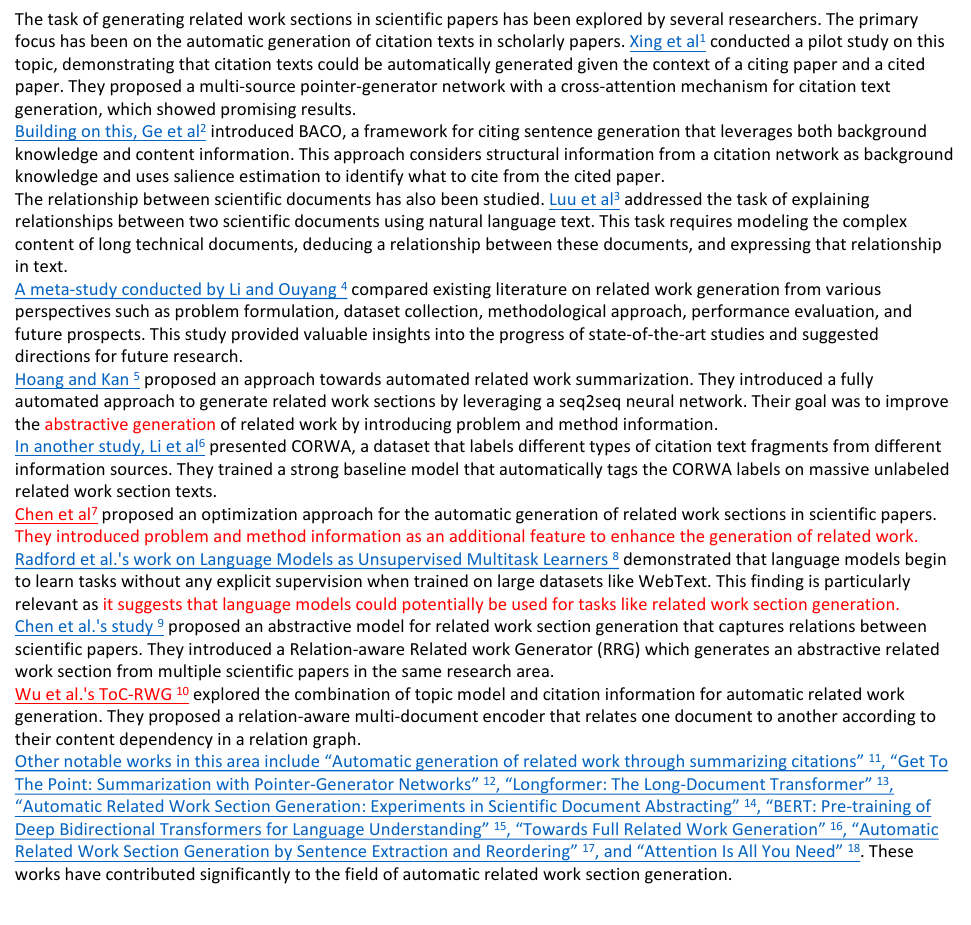}
         \vspace{-2.0em}
         \caption{Bing Chat output (``more precise'' setting).}
         \label{fig:bing_chat}
     \end{subfigure}
     \quad
     \begin{subfigure}[b]{0.48\textwidth}
         \centering
         \includegraphics[width=\textwidth]{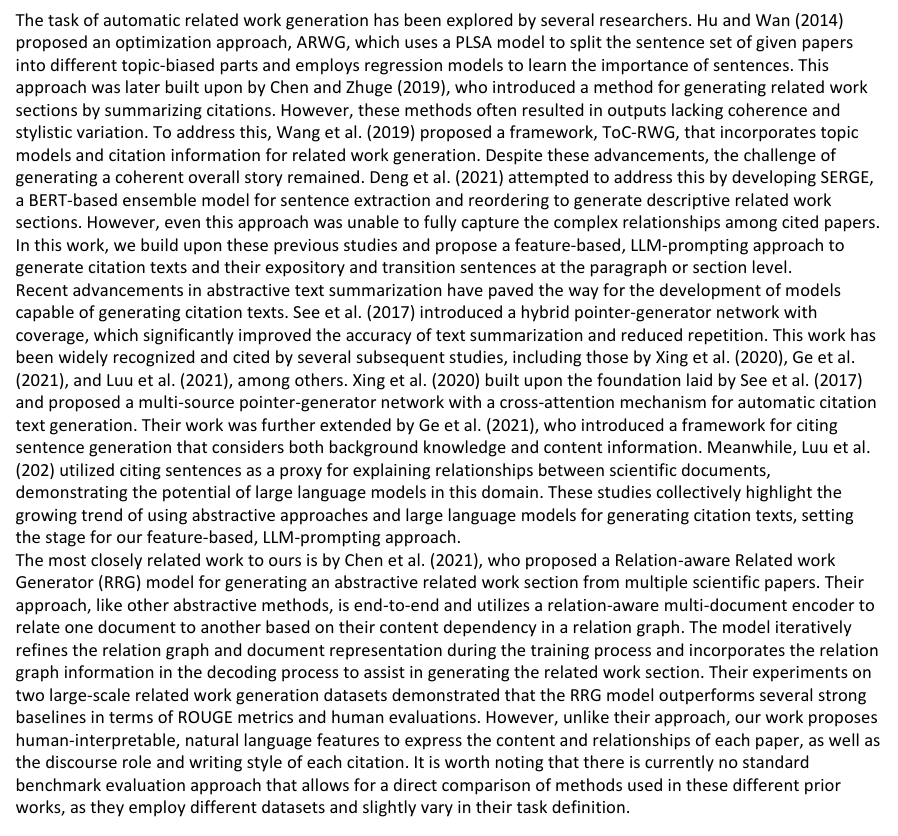}
         \vspace{-1.5em}
         \caption{Our approach.}
         \label{fig:ours}
     \end{subfigure}
  
     \vspace{-0.5em}
        \caption{Comparison between GPT4-powered Bing Chat and our approach on reproducing Section \ref{sec:related_work} of this paper. Bing Chat is given a prompt consisting of the title and abstract of this paper, as well as a list of reference paper titles, which fits comfortably in its context window. Bing Chat's output is generic, ill-organized, and non-factual; statements in \textcolor{red}{red} are misattributed or incorrect.}
        \label{fig:comparison}
    \vspace{-2.0em}
\end{figure*}

\section{Introduction} \label{sec:introducton}

Due to the rapid pace of research publications, including pre-prints that have not yet been peer-reviewed, keeping up to date with the latest work is very time-consuming. Even with daily feed tools like the Semantic Scholar Research Feed\footnote{\url{https://www.semanticscholar.org/faq/what-are-research-feeds}}, researchers must still curate, read, and digest all the new papers in their feed. Thus, there is a need for concise, automatically generated literature reviews summarizing the set of new papers in a feed, customized for the researcher whose feed it is. 

Unfortunately, there does not exist a dataset of such literature reviews. Survey articles are similar, but they are very long, not customized to a specific reader, and relatively rare. In this work, we use the related work sections of scientific articles as a proxy for the kind of short, customized, daily feed summaries we wish to generate. Related work sections have several advantages: they are concise, usually no more than one page long; they are customized to their parent article, just as a daily feed summary should be customized to its owner; and they are plentiful and can be automatically extracted as the ``Related Work," ``Literature Review," or ``Introduction" sections of scientific articles. To relate back to a daily feed summary, one can easily imagine using the most similar paper authored by the feed owner as the ``citing" paper perspective from which to customize the literature review.


The task of automatically generating citations for scientific articles has been explored using both extractive \cite{hoang-kan-2010-towards, hu-wan-2014-automatic, chen2019automatic, wang2019toc, deng2021automatic} and abstractive techniques \cite{abura2020automatic, xing-etal-2020-automatic, ge-etal-2021-baco, luu-etal-2021-explaining, chen-etal-2021-capturing, li-etal-2022-corwa}. However, we argue that the dominant approach of generating a single citation sentence in isolation ignores the relationships \textit{among} cited papers, which are just as important as that between the citing and cited papers. Literature reviews, whether for a related work section or a daily feed summary, are multi-document summaries and should contain the non-citation, expository and transition sentences needed to compose a coherent story \cite{li-etal-2022-corwa}. Recent neural approaches frame citation generation as an end-to-end, sequence-to-sequence task; they are thus constrained by the length limitations of their models --- research papers are long documents --- and are unable to make use of supporting features, such as citation intents or topic information, which would require training additional models.

The recent dramatic success of prompting-based methods using LLMs, like Chat-GPT and GPT-4 \cite{openai2023gpt4}, makes it possible to pursue a feature-based approach to generating richer citation texts, as well as generating multiple citations at a time to better capture the complex relationships among research papers. However, as Figure \ref{fig:comparison} shows, even a SOTA LLM, augmented with a search engine to retrieve papers \footnote{\url{https://www.bing.com/new}}, cannot generate a factually-correct, on-topic literature review from scratch. Bing Chat with GPT-4 hallucinates cited papers' approaches, and the expository sentences are generic and vague. We can see that the LLM needs guidance in identifying the relevant contributions of each cited paper
, as well as how the cited papers relate to one another.  

In this work, as a first step towards generating customized daily feed summaries, we explore a feature-based, LLM-prompting approach to generating citation texts and their transition sentences at the paragraph level, using automatically extracted related work sections as the evaluation targets. Our main contributions are as follows:
\begin{itemize}
    \item We propose features capturing the relationships between cited and citing papers and among papers cited together. We show that these features can be extracted by prompting LLMs and compose them into a new prompt for generating several citations, along with transition sentences, in one pass.
    \item We conduct experiments on a planning-based setting, where a plan consisting of a few sentences describing the high-level relationships among cited papers is used to guide generation. In this preliminary study, we use a human-provided plan to investigate the impact of these guiding ``main ideas" on the organization of citations in the generated paragraphs.
    \item We perform an expert evaluation to investigate the impact of our proposed features on the quality of the generated paragraphs. We find a strong correlation between human preference and integrative writing style, suggesting that readers prefer more high-level, abstract citations, with transition sentences between them to provide a coherent overall story. 
\end{itemize}

\section{Background and Related Work} \label{sec:related_work}

\citet{hoang-kan-2010-towards} proposed the task of automatic related work generation: generating the related work section of a target paper given a list of papers to cite, assuming the rest part of the target paper is available. Early extractive approaches automatically select and concatenate salient sentences from the cited papers \cite{hu-wan-2014-automatic, chen2019automatic, wang2019toc, deng2021automatic}. 
As a result, their outputs lack coherence among citations and have no overall story, and the sentences lack stylistic variation; transitions and sentences relating back to the target paper are impossible to produce using an extractive approach.

More recently, abstractive approaches have focused on generating a single citation at a time, given the cited paper and assuming the rest of target paper, including the rest of the related work section, is available \cite{abura2020automatic, xing-etal-2020-automatic, ge-etal-2021-baco, luu-etal-2021-explaining, li-etal-2022-corwa}. These works used a variety of architectures (pointer-generator \cite{see-etal-2017-get}, vanilla Transformer \cite{vaswani2017attention}, GPT-2 \cite{radford2019language} or Longformer-Encoder-Decoder \citep[LED;][]{beltagy2020longformer}) to generate citations from cited paper abstracts; the cited paper full texts were not used due to their length. 

The most similar work to ours is \citet{chen-etal-2021-capturing}, who attempt to generate multiple citations at once. However, their approach, like that of other abstractive prior works, is end-to-end; they augment a document encoder with a graph network to learn relationships among document representations. In contrast, we propose human-interpretable, natural language features to express paper content and relationships, as well as the discourse role and writing style of each citation.

Finally, as \citet{li2022automatic} note, there is no standard benchmark to directly compare methods from these prior works, which use different datasets and vary slightly on the task definition.

\section{Approach}
\label{sec:approach}

We use multi-stage prompting of LLMs to first collect features from each cited paper and then compose them into a prompt for generating a paragraph of citations with supporting expository and transition sentences. It is well-known that, given a simple prompt, such as ``Generate a literature review about XXX," LLMs produce poorly-organized and inaccurate citations; from our Bing Chat example, we can see that the LLM 
inaccurately describes the approaches and contributions of the cited papers. We must provide the LLM with more detailed and specific information about the cited papers. 

We identify two sets of key support features. First, we extract features for the target paper and each cited paper from a local citation network, capturing information about the relationship between the target and each cited paper, as well as between pairs of cited papers (Section \ref{sec:cited_features}). Second, we extract features from the text of the target paper itself, which provide contextual information to ensure the generated citations stay on topic (Section \ref{sec:citing_features}); in our experiments, the target paper stands in for the daily feed owner so that the literature review is customized for their perspective.

After extracting these features, we compose a prompt to generate paragraphs\footnote{The number of paragraphs we generate in one shot depends on the total number of cited papers; if there are too many, the input prompt becomes too long to generate more than one paragraph at a time.} of citations (Section \ref{sec:full_related_work_generation}). Finally, we use the generated draft to extract candidate cited text spans for each cited paper, add them to the prompt, and re-generate improved paragraphs (Section \ref{sec:CTS}). Our prompts are shown in Appendix Tables \ref{tab:faceted_summary_prompt}--\ref{tab:full_related_work_generation_prompt}; Appendices \ref{sec:examples} \& \ref{sec:example_related_work} show examples of features and paragraphs.

\subsection{Citation Network Features} \label{sec:citation_network_features}
\label{sec:cited_features}
We identify key features under the framework of a local citation network centered on the target paper. Each node represents a paper, and an edge represents the relationship between two papers. Unlike \citet{chen-etal-2021-capturing}, we use \emph{natural language descriptions} as network features, allowing us to leverage the seq2seq nature of LLMs, rather using numerical feature vectors with a graph neural network; the natural language descriptions also improve the interpretability of the citation network. 

\paragraph{Faceted summary.} Each node in the citation network represents a paper, and its core feature is the faceted summary \cite{meng-etal-2021-bringing}, which highlights the key aspects of the paper for rapid understanding: the paper's \emph{objective}, \emph{method}, \emph{findings}, \emph{contributions} and \emph{keywords}. Just as a human may quickly skim the title, abstract, introduction, or conclusion (TAIC) sections to get the gist of a paper, we focus on the most important facets of a paper when generating its citation. The faceted summary also provides the practical benefit of reducing the number of tokens needed to represent a paper; the limited input window size of LLMs encourages a compact representation for each node. We prompt the LLM to generate a faceted summary given the TAIC of each paper.

\paragraph{Relationship between paper pairs.} Each edge in the citation network represents the relationship between the two paper nodes it connects. Given a pair of papers $A$ and $B$, we leverage the LLM's strong summarization ability to synthesize information from all citation spans\footnote{We use \citet{li-etal-2022-corwa}'s citation tagger to extract spans.} where paper $A$ cites paper $B$, conditioned on the faceted summaries of $A$ and $B$, into a single natural language description of their (directed) relationship. 
The incoming edges on paper $B$'s node thus capture how $B$ has been discussed by other works, providing a history of how its ideas have influenced its field; the outgoing edges from likewise capture how $B$ has developed ideas from other works in its field.

\paragraph{Enriched citation intent \& usage.} 
Citation intents encode why and how an author cites a paper: to give \textit{background} information, to use a proposed \textit{methodology}, or to \textit{compare} experimental results. Existing work on citation intent focuses on proposing new label sets or modeling approaches to predict those labels, without applying them to downstream applications \cite{garfield1965can, teufel-etal-2006-automatic, dong-schafer-2011-ensemble, jurgens-etal-2018-measuring, cohan-etal-2019-structural, tuarob2019automatic, zhao-etal-2019-context}, and they framed intent prediction as a classification task to reduce its complexity. However, as \citet{lauscher-etal-2022-multicite} point out, simple classification label sets struggle to represent ambiguous, real-world citations. To the best of our knowledge, we are the first to apply citation intent to citation generation and to extend intent labels to rich natural language descriptions. 

In addition, \citet{li-etal-2022-corwa} distinguish between \emph{dominant}- and \emph{reference}-type citations. For example, consider the sentence ``\citet{luu-etal-2021-explaining} fine-tuned GPT-2 \cite{radford2019language} to predict citation sentences." The emphasis is on the \emph{dominant}-type citation of Luu et al., whereas the \emph{reference}-type citation of Radford et al. is not explained in detail, since GPT-2 is being cited as a tool. Ignoring this distinction results in unnatural-sounding paragraphs that treat all cited papers as equally-important, \emph{dominant}-type citations. 

Thus, for each cited paper $B$, we prompt the LLM to summarize how other papers $A_i$ in the network cite $B$ (intent) as well as if the majority usage of $B$ is as an important, \emph{dominant} cited paper or if it is simply cited for \emph{reference}. The prompt includes the faceted summaries (node features) of the other citing papers $A_i$, all relationships (edge features) between $A_i$ and $B$, and the text of the citation spans for $B$ in $A_i$. 
This enriched citation intent/usage feature roughly corresponds to a discursive summary of all edges incident to node $B$. 

\subsection{Target Paper Features} \label{sec:citing_features}

To generate paragraphs with a coherent overall story, we collect features from the target paper, capturing the context and perspective of the reader.

\paragraph{Title, abstract, introduction, and conclusion (TAIC).} Despite the powerful zero-shot generation ability of LLMs, they are typically not trained specifically for scientific document generation and lack the necessary domain knowledge to write like a domain expert. 
We leverage the LLM's strong in-context learning ability by including the full text of the TAIC sections of the target paper. The TAIC provides context to the LLM, so that the story and tone of the TAIC can inform the focus and organization of the citations to be generated; in our experiments, the target paper represents the reader's interests, so a good literature review should be coherent with the target paper. 

\paragraph{Guiding plan of main ideas.} Intuitively, there can be multiple plausible literature reviews for the same set of cited papers. A reader may prefer one over another, even though they are all factually correct, depending on the perspective given by the target paper. To better capture this information, we experiment with a human-provided plan to guide generation; we leave the automatic generation of a guiding plan to future work. The plan is a short summary of the main ideas to be discussed; to simulate this feature during evaluation, we prompt an LLM to condense the gold related work section of the target paper into a short summary of its main ideas, \textit{ignoring citations} to avoid information leak.

\subsection{Related Work Paragraph Generation}
\label{sec:full_related_work_generation}
\begin{figure}[t]
\small
\begin{center}
\setlength{\tabcolsep}{5pt} 
    \begin{tabular}{p{0.8\linewidth} }
    \hline
    \textbf{Prompt} \\ \hline
    The title, abstract, introduction and conclusion section of the target paper are as follows: \\
    Title: \{\{title\}\} \\
    Abstract: \{\{abstract\}\} \\
    Introduction: \{\{introduction\}\} \\
    Conclusion: \{\{conclusion\}\} \\
    ...
Write a literature review that concisely cites the following papers in a natural way using all of the main ideas as the main story.
...
You can freely reorder the cited papers to adapt to the main ideas.
...  \\ \hdashline
  Main idea of our literature review: \\ 
  \{\{\emph{main ideas}\}\} \\ \hdashline
    List of cited papers: \\
    1. \{\{titleB1\}\} by \{\{authorB1\}\} et al. \{\{yearB1\}\} \\
    \{\{\emph{Faceted Summary or Abstract of B1}\}\} \\ 
    <Usage> \{\{Enriched citation usage of B1\}\} \\
    How other papers cite it: \\
    \{\{Relation between Ax and B1\}\} \\
    \{\{Relation between Ay and B1\}\} \\
    ... \\
    \hdashline
    Potentially useful sentences from the target paper: 
    \{\{section \#1\}\} \{\{CTS \#1\}\} \\
    \{\{section \#2\}\} \{\{CTS \#2\}\} \\
    ... \\ \hdashline
    2. \{\{titleB2\}\} by \{\{authorB2\}\} et al. \{\{yearB2\}\} \\
    ...... \\
    \hline
    \end{tabular}
    \vspace{-0.5em}
    \caption{Prompt format for generating a literature review paragraph (simplified for length). 
    } \label{fig:related_work_generation_prompt}
    \vspace{-1.5em}
\end{center}
\end{figure}

With the features above, we prompt the LLM to generate one paragraph, subsection, or when length allows, the entire literature review in one pass (Figure \ref{fig:related_work_generation_prompt}). We find section-level generation to be the most robust, as the LLM often does not follow prompt instructions as closely when generating paragraph-by-paragraph, but we are constrained by the length limit of the LLM. During evaluation, for target related work sections with many cited papers, such that the full prompt exceeds the LLM's limit, we manually chunk the gold related work section based on subsections or titled paragraphs, partition the cited papers according to those chunks, and generate each chunk individually, concatenating the generated subsections or paragraphs in the same order as in the gold related work section.

\subsection{Enhancing Details with CTS} \label{sec:CTS}

We observe that the generated citations may lack detail compared to human-written ones. This makes sense because the generation prompt contains only summaries of, but no actual text from, each cited paper. To supply more detail, we follow \citet{yasunaga2019scisummnet, wang2019toc} in using ROUGE-based ranking \cite{cao2015ranking} to retrieve \textit{cited text spans} \citep[CTS;][]{jaidka2018insights, jaidka2019cl, aburaed-etal-2020-multi}: the cited paper text span most relevant to the corresponding citation.

We retrieve CTS using a newly generated candidate citation span as query, which we extract using \citet{li-etal-2022-corwa}'s citation tagger; we compute the average of ROUGE-1 and -2 recall scores against each sentence in the corresponding cited paper. We take the top-$k$\footnote{We adjust $k$ case-by-case so the prompt length does not exceed the LLM's input window, with a hard cap at $k=10$.} sentences as CTS to augment the prompt and then re-generate the paragraph. 

\section{Experimental Settings}
For each target related work section, we start with the PDF of the target paper and aim to generate a plausible candidate to replace the gold related work section using our feature-based approach. 

As we discuss in Sections \ref{sec:introducton}--\ref{sec:related_work}, to the best of our knowledge, there is no prior baseline that can generate the long texts of this task; even a SOTA LLM, Bing Chat with GPT-4, fails to be a competitive baseline. Therefore, we focus on comparing different input feature variations (Section \ref{sec:variants}).

\subsection{Implementation Details} \label{sec:implementation_details}
We use Google search API \footnote{\url{https://pypi.org/project/googlesearch-python/}} to find and download cited papers and doc2json \cite{lo-etal-2020-s2orc} to parse PDFs into JSON format. We use Chat-GPT (gpt-3.5-turbo-0301) for all feature extraction steps and GPT-4 (gpt-4-0314; maximum input length limit of 8k tokens) for the generation step. 

We do not perform any training or fine-tuning, and we do not use any datasets; we only use the pre-trained citation tagger of \citet{li-etal-2022-corwa} for citation span extraction. We design our prompts using the first author's previous publications as a development set and use our human judges' nominated papers as our test set. 

\subsection{Human Evaluation Settings} 
Because it is challenging for non-domain experts to evaluate the quality and factuality of scientific texts, we conduct a human evaluation by inviting domain experts to evaluate candidate literature reviews generated for one of their own published papers, or a paper closely related to their own work. Our experts are all fluent in English and are a mix of Ph.D. students and post-doctorate researchers in both academia and industry. We instruct them to nominate a target paper such that they are very familiar with all of the cited papers.

Because the experts were recruited from among our colleagues, about half of the papers evaluated are natural language processing papers, with the other half from other computational fields, including machine learning, speech processing, computer vision, robotics, computer graphics, and programming languages. Almost all papers were published after September 2021 and so were not included in the training data of the LLMs (Appendix Table \ref{tab:year}).

The evaluators are asked to score each generated related work section in terms of (1) \emph{fluency}, (2) \emph{organization \& coherence}, (3) \emph{relevance} to the target paper, (4) \emph{relevance} to the cited papers, (5) \emph{factuality} and the number of non-factual or inaccurate statements, (6) \emph{usefulness \& informativeness}, (7) \emph{writing style}, and (8) \emph{overall quality}.

\subsection{Input Feature Variations} \label{sec:variants}

\begin{table}[t]
\small
\begin{center}
\setlength{\tabcolsep}{2.5pt} 
\renewcommand{\arraystretch}{1.0} 
    \begin{tabular}{lllllllll}
    \hline
    \textbf{Feature} & \textbf{A} & \textbf{B} & \textbf{C} & \textbf{D} & \textbf{E} & \textbf{F} & \textbf{G} & \textbf{H} \\ 
    \hline
    main idea & \cmark & - & \cmark & \cmark & \cmark & \cmark & \cmark & \cmark \\ \hline
    target TAIC & \cmark & \cmark & - & \cmark & \cmark & \cmark & \cmark & \cmark \\ \hline
    faceted summary & \cmark & \cmark & \cmark & - & \cmark & \cmark & \cmark & \cmark \\ 
    cited abstract & - & - & - & \cmark & - & - & - & - \\ \hline
    intent/usage & \cmark & \cmark & \cmark & \cmark & - & \cmark & - & \cmark \\
    relationship & \cmark & \cmark & \cmark & \cmark & \cmark & - & - & \cmark \\ \hline
    CTS & - & - & - & - & - & - & - & \cmark \\ \hline

    \end{tabular}
    \vspace{-0.5em}
    \caption{Features in each generation variant.} \label{tab:configurations}
    \vspace{-1.5em}
\end{center}
\end{table}

Table \ref{tab:configurations} shows the input features used by each of our generated variants. $A$ is our baseline, with access to all features, including the human-provided main idea plan. $B$ is the only variant that does not use the main ideas, making it the only variant that could be generated completely automatically. Variant $C$ ablates the TAIC of the target paper, while $D$ replaces each cited paper's faceted summary with its abstract. Variants $E$, $F$, and $G$ ablate the enriched citation intent and usage, the relationship between paper pairs, and both, respectively. Finally, variant $H$ adds the CTS-based re-generation step. 

\section{Results and Analyses}

\begin{table}[t]
\small
\begin{center}
\setlength{\tabcolsep}{2.5pt} 
\renewcommand{\arraystretch}{1.0} 
    \begin{tabular}{llll}
    \hline
    \textbf{Variant} & \textbf{ROUGE-1} & \textbf{ROUGE-2} & \textbf{ROUGE-L} \\ \hline
     A & 0.513 & 0.216 & 0.248 \\ \hline
     B & \textit{0.446} & \textit{0.131} & \textit{0.177} \\ \hline
     C & \textit{0.501} & \textit{0.201} & \textit{0.235} \\ \hline
     D & \textit{0.511} & \textit{0.215} & \textbf{0.249} \\ \hline
     E & \textbf{0.514} & \textbf{0.223} & \textbf{0.255} \\ \hline
     F & \textbf{0.520} & \textbf{0.221} & \textbf{0.252} \\ \hline
     G & \textbf{0.517} & \textbf{0.225} & \textbf{0.256} \\ \hline
     H & 0.513 & \textit{0.215} & \textbf{0.249} \\ \hline
    \end{tabular}
    \vspace{-0.5em}
    \caption{ROUGE scores of generated variants evaluated against the gold related work sections. \textbf{Bold} indicates improvement over baseline $A$, while \textit{italics} indicate lowered performance.} \label{tab:rouge}
    \vspace{-1.5em}
\end{center}
\end{table}

\newcommand{\tablerow}[9]{ #1 & #3 & #2 & #6 & #8 & #4 & #5 & #9 & #7}
\begin{table}[t]
\small
\begin{center}
\setlength{\tabcolsep}{2pt} 
\renewcommand{\arraystretch}{1.0} 
    \begin{tabular}{lllllllll}
    \hline
    \textbf{Metrics} & \textbf{A} & \textbf{B} & \textbf{C} & \textbf{D} & \textbf{E} & \textbf{F} & \textbf{G} & \textbf{H} \\ \hline
    \tablerow{Fluency}{$3.78$}{$4.11$}{$4.07$}{$4.11$}{$4.00$}{$4.19$}{$\mathbf{4.33}$}{$4.15$} \\ \hline
    \tablerow{Coherence}{$3.07$}{$3.30$}{$3.59$}{$3.59$}{$3.33$}{$3.37$}{$\mathbf{3.70}$}{$3.52$} \\ \hline
    \tablerow{Rel target}{$3.67$}{$3.78$}{$4.11$}{$4.00$}{$3.89$}{$4.00$}{$\mathbf{4.19}$}{$4.07$} \\ \hline
    \tablerow{Rel cited}{$3.93$}{$\mathbf{4.22}$}{$4.19$}{$4.19$}{$4.15$}{$4.04$}{$\mathbf{4.22}$}{$4.00$} \\ \hline
    \tablerow{Factuality}{$3.89$}{$4.04$}{$3.93$}{$\mathbf{4.30}$}{$3.74$}{$3.74$}{$3.89$}{$3.93$} \\ \hline
    \tablerow{Usefulness}{$3.30$}{$3.74$}{$\mathbf{3.85}$}{$3.70$}{$3.59$}{$3.78$}{$3.52$}{$3.59$} \\ \hline
    \tablerow{Writing}{$3.07$}{$3.48$}{$3.70$}{$3.52$}{$3.30$}{$\mathbf{3.81}$}{$3.78$}{$3.44$} \\ \hline
    \tablerow{Overall}{$2.89$}{$3.33$}{$\mathbf{3.67}$}{$3.56$}{$3.15$}{$3.15$}{$3.48$}{$3.22$} \\ \hline
    \tablerow{\# error}{$0.78$}{$0.70$}{$0.81$}{$\mathbf{0.44}$}{$0.78$}{$0.89$}{$0.70$}{$0.74$} \\ \hline
    \end{tabular}
    \vspace{-0.5em}
    \caption{Average human evaluation scores.} \label{tab:human_evaluation}
    \vspace{-1.5em}
\end{center}
\end{table}

\paragraph{Automatic Evaluation.} Table \ref{tab:rouge} shows the ROUGE scores of our generated variants compared to the gold related work sections. Overall, most variants yield decent scores, indicating that they are mostly on-topic. Notably, variant $B$ has significantly lower ROUGE scores than other variants, which makes sense because it is the only one without the main idea plan. This emphasizes the important and irreplaceable nature of the guiding plan. Variant $C$, which has the main ideas but no target paper TAIC is the next lowest, again suggesting that features related to the reader's perspective are the most important for a good literature review.

\begin{table}[t]
\small
\begin{center}
\setlength{\tabcolsep}{3pt} 
\renewcommand{\arraystretch}{1.0} 
    \begin{tabular}{cccccc}
    \hline
    \textbf{Diff} & \textbf{Vrt}&  \textbf{\heart Vrt\%} & \textbf{Tie\%} & \textbf{\heart Bsl\%} \\ \hline
    $-$ main story  & B & 22.2  & 29.6& \textbf{48.1}\\ \hline
    $-$ target TAIC  & C  & 22.2 & 37.0& \textbf{40.7}\\ \hline
    $-$ faceted summary  & \multirow{2}{*}{D}  & \multirow{2}{*}{29.6}& \multirow{2}{*}{\textbf{48.1}} & \multirow{2}{*}{22.2} \\ 
    $+$ cited abstract & & & & \\ \hline
    $-$ intent/usage  & E  & \textbf{40.7}& 37.0& 22.2  \\ \hline
    $-$ relationship  & F  & \textbf{40.7} & 29.6& 29.6\\ \hline
    $-$ intent/usage  & \multirow{2}{*}{G} & \multirow{2}{*}{22.2} & \multirow{2}{*}{33.3} & \multirow{2}{*}{\textbf{44.4}}\\
    $-$ relationship & & & & \\ \hline
    $+$ CTS  & H  & 25.9& \textbf{37.0}&  \textbf{37.0}\\ \hline
    \end{tabular}
    \vspace{-0.5em}
    \caption{Comparison of human \textit{overall} scores across variants, with respect to the baseline $A$.} \label{tab:human_evaluation_comparison}
    \vspace{-1.5em}
\end{center}
\end{table}

\paragraph{Human Evaluation.} Due to the challenging and expensive nature of evaluating highly specialized academic research papers, we are only able to evaluate one target related work section per domain expert judge, with 27 judges in total. Table \ref{tab:human_evaluation} shows the average human evaluation scores across all 27 judges. Writing is a highly personal and idiosyncratic process, and the high variance in the human evaluation scores reflects this fact; 
different variants are preferred by different judges. 

\subsection{Importance of Input Features} 
We integrate the results of Tables \ref{tab:rouge}, \ref{tab:human_evaluation} \& \ref{tab:human_evaluation_comparison} to analyze the usefulness of each input feature.

\paragraph{Main idea plan.} All tables show that baseline $A$ outperforms variant $B$, which ablates the main idea feature. The fact that main idea information is not found in any other feature (Appendix \ref{sec:component_analysis}) confirms the importance of a human-provided main idea to guide the LLM in generating a satisfactory story.

\paragraph{Target paper TAIC.} The baseline $A$ also outperforms variant $C$, which ablates the target paper title, abstract, introduction, and conclusion, confirming our hypothesis that this feature provides crucial context for the generated paragraphs.

\paragraph{Enriched citation usage \& relationship between papers.} Comparing $A$ to variants $E$, $F$, and $G$, we find a weak trend that access to either enriched citation usage or relationship between papers is helpful, with the latter slightly preferred; this finding is consistent with the observed \textit{coverage} and \textit{density} discussed in Appendix \ref{sec:component_analysis}. Variants with access to both or neither feature underperform, suggesting that, while the usage and relationship features are important, they are also mutually redundant (the usage feature of a cited paper summarizes all its relationships) and the presence of both causes the LLM to over-emphasize this information.

\paragraph{Faceted summaries \& cited paper abstracts.} We are surprised to observe that variant $D$, which uses cited paper abstracts instead of faceted summaries, outperforms $A$ in terms of \textit{fluency}, \textit{coherence}, \textit{writing}, and \textit{overall}. We note that $D$ still has indirect access to the faceted summaries since the relationship-between-papers feature is derived from faceted summaries. We hypothesize that $D$ is preferred over $A$ because abstracts are human-written and contain more narrative-style sentences than the LLM-generated faceted summaries (see Section \ref{sec:corwa_analysis}). However, the baseline $A$ still outperforms $D$ in terms of \textit{informativeness} and \textit{factuality}. 

\paragraph{CTS.} Comparing the baseline $A$ with variant $H$, we observe that re-generating the related work section using cited text spans is a controversial choice that leads to very high variance in human evaluation scores. 44\% of the judges report improved \textit{writing} style, and 26\% report improved \textit{informativeness}, while 30\% report decreased \textit{factuality}. 
Therefore, CTS should be used cautiously. 

\subsection{Analysis of Writing Style} \label{sec:corwa_analysis}

\begin{table}[t]
\small
\begin{center}
\setlength{\tabcolsep}{1.5pt} 
\renewcommand{\arraystretch}{1.0} 
    \begin{tabular}{c|c|cccccccc}
    \hline
    \textbf{Label\%} & \textbf{Gld} & \textbf{A} & \textbf{B} & \textbf{C} & \textbf{D} & \textbf{E} & \textbf{F} & \textbf{G} & \textbf{H} \\ \hline
    Transition &  31.1 &  \textit{17.0} &  \textit{10.9} &  \textit{18.5} &  \textit{19.7} &  \textit{17.0} &  \textit{19.3} &  \textit{19.6} &  \textit{20.1} \\ \hline
    Single-Sum &  28.2 &  \textbf{47.2} &  \textbf{59.8} &  \textbf{51.7} &  \textbf{40.7} &  \textbf{45.7} &  \textbf{46.2} &  \textbf{45.4} &  \textbf{40.7} \\ \hline
    Narrative &  20.8 &  \textit{11.3} &  \textit{3.5} &  \textit{8.4} &  \textit{13.5} &  \textit{14.5} &  \textit{10.0} &  \textit{14.1} &  \textit{14.1} \\ \hline
    Reflection &  15.4 &  \textbf{17.0} &  \textbf{21.6} &  15.4 &  \textbf{19.3} &  \textbf{17.4} &  \textbf{16.9} &  \textbf{16.9} &  \textbf{17.8} \\ \hline
    Multi-Sum &  3.6 &  \textbf{7.5} &  \textit{3.3} &  \textbf{6.0} &  \textbf{6.8} &  \textbf{5.5} &  \textbf{7.4} &  \textbf{4.0} &  \textbf{6.6} \\ \hline \hline
    Dominant &  34.7 &  \textbf{70.0} &  \textbf{81.4} &  \textbf{77.1} &  \textbf{64.7} &  \textbf{63.6} &  \textbf{70.3} &  \textbf{60.1} &  \textbf{63.2} \\ \hline
    Reference &  65.3 &  \textit{30.0} &  \textit{18.6} &  \textit{22.9} &  \textit{35.3} &  \textit{36.4} &  \textit{29.7} &  \textit{39.9} &  \textit{36.8} \\ \hline
    \end{tabular}
    \vspace{-0.5em}
    \caption{\citet{li-etal-2022-corwa} writing style analysis. Percentage of the discourse role of sentences (top) or citation types (bottom) within each variant. \textbf{Bold} indicates styles used more frequently in generated variants than gold related work sections; \textit{italics} indicate less frequent styles.} \label{tab:corwa_analysis}
\end{center}
\end{table}


\citet{khoo2011analysis} studied the writing style of literature reviews by categorizing them as \textit{integrative} or \textit{descriptive}, depending on whether they focus on high-level ideas or on detailed information from specific studies. \citet{li-etal-2022-corwa} extended this distinction to individual citations, distinguishing \emph{dominant} citations, which focus on and describe cited papers in detail, and \emph{reference} citations, which are short, highly abstracted, and often tangential to the rest of the sentence. Li et al. also introduced sentence-level discourse roles: \textit{transition} and \textit{narrative} sentences provide exposition and high-level observations; \textit{single} and \textit{multi-summarization} sentences give specific, detailed information about one or more cited papers; and \textit{reflection} sentences relate cited papers to the target paper.

To study the writing style of our generated paragraphs, we use Li et al.'s citation tagger to label the usage types and discourse roles of the gold related work sentences and our variants. As Table \ref{tab:corwa_analysis} shows, there is a huge gap between the two writing styles: gold sentences mainly consist of transition and narrative sentences with reference-type citations, while all generated variants have far more explicit single-summary sentences with dominant-type citations. In other words, the generated paragraphs are mostly \textit{descriptive}, consisting of individual paper summaries, rather than a coherent story \textit{integrating} all the cited papers.

\subsection{Correlation Among Human Preference, ROUGE, and Writing Style} \label{sec:correlation}
As Tables \ref{tab:rouge}, \ref{tab:human_evaluation} \& \ref{tab:corwa_analysis} show, there is a strong correlation between human preference (\textit{overall} score) and ROUGE-L scores, as well as between ROUGE scores and writing style (proportion of reference-type citations), with Kendall's $\tau$ of 0.592 and 0.691, respectively. This suggests that we can use automatic metrics such as ROUGE and the proportion of reference-type citations to estimate human judgments, which is extremely challenging to collect on a large scale. Moreover, this observation emphasizes the importance of having a coherent and organized story consisting of narrative-style sentences with reference-type citations and transition sentences bridging between them. 

\subsection{Qualitative Error Analysis}
Despite the overall success of our approach --- over half the judges wrote that the generated variants would be good first drafts for the gold related work sections --- our collected comments from the judges show that composing a literature review is still a very challenging task. We summarize the typical issues mentioned by the judges below:

\paragraph{Factual errors.} While all generated variants have a small absolute number of factual errors (see Table \ref{tab:human_evaluation}), incorrect statements are the most frequently mentioned problem. 
For example, one judge complained, ``\ldots two descriptions are false (Hearst patterns not extracted from Wikipedia, and there were no edits in Bowman et al.).'' We observe that the overall human evaluation score correlates with the factuality score (Kendall's $\tau$=0.50).

\paragraph{Emphasizing the right cited papers.} 
A good literature review should have a logical story; simply concatenating individual cited paper summaries is not sufficient. Our judges complained about less important papers receiving too much attention:
\begin{itemize}[noitemsep, nolistsep]
    \item ``\ldots too much detail for the papers and has a paragraph on human-in-the-loop data generation which is not very relevant to the paper (should be mentioned briefly).''
    \item ``The descriptions of the cited papers, while accurate, seem less relevant to the citing paper and to the story as a whole, and the papers that get a lengthier description are not the most central ones.''
    \item ``The focus on traditional decompilation methods is too strong for the paper content.''
\end{itemize}

Further, due to the nature of using related work sections as our evaluation targets, the publication dates of the cited papers can vary greatly. While this would not be a problem in a daily feed literature review (since all of the papers would be new), our judges were more likely to complain if the generated literature review focused to much on earlier works: ``\ldots why does it focus on approaches from 20 years ago than recent approaches?" These comments confirm our finding that dominant, summarization-type citation sentences may not be appropriate for all cited papers.

\paragraph{Paragraph organization.} How to group similar works together is a major challenge. While our approach is usually able to generate well-organized texts when the variant includes sufficient features, failure cases significantly impact the human evaluation. For example, one judge wrote, ``Second paragraph starts with `In the pursuit of automating reinforcement learning', but then immediately cite Henderson et al. (2018) which talks about reproducibility and not automating RL issues.''

Judges also commented that they would prefer more comparisons among cited papers: ``\ldots some citations did not highlight their difference from other work, such as Schick et al. (2021) generates pairs of data.''

\paragraph{} Other observed issues include insufficient evidence for claims and inconsistent citation formatting. 
In addition, the LLM does not always follow the prompts, occasionally resulting in some cited papers being silently dropped from the output.

Overall, we find that organization and flow between citations is very important to the judges and is mentioned in most positive comments:
\begin{itemize}[noitemsep, nolistsep]
    \item ``Good connection between the explained works. It is not just a list of contributions.''
    \item ``The organization is close to perfect, and the story flows well in this one. One citation is missing, and, surprisingly, one citation was added (Kondadadi, 2013) - in exactly the right place and with an accurate description! weird but pretty cool!''
\end{itemize}


\section{Conclusion} \label{sec:conclusion}

We have presented a feature-based approach for prompting LLMs to explain the relationships among cited papers. With the ultimate goal of generating a literature review summarizing the contents of a researcher's daily paper feed, we have conducted a pilot study using the related work sections of scientific articles as a proxy for the kind of literature reviews we wish to generate: short, customized for a particular target paper (standing in for daily feed's owner), and focused on explaining how the cited papers relate to each other and why they are important. Our approach focuses on using the strong natural language understanding and summarization abilities of LLMs to extract interpretable, natural language features describing the content of the cited and target papers, as well as their relationships with each other and with other papers that have cited them in the past. We also propose a ``main ideas" plan to guide the LLM to generate a coherent story, using a human-supplied plan in these preliminary experiments.

Our detailed expert evaluation reveals that human judges dislike literature reviews that simply concatenate cited paper summaries together, demonstrating the importance of generating at the paragraph or section level, including transition sentences, rather than focusing on individual citations. Human judges are also sensitive to the relevance of each cited paper and strongly dislike generations that wrongly emphasize less impactful papers. We conclude that accurate descriptions of a cited paper's methodology are not the only important facet of scientific document processing --- understanding the rich and sophisticated relationships among papers is the key.

\section*{Limitations}

\paragraph{Citation retrieval.} In their written comments, several judges expressed the wish that our system would help them find other related papers to read. This is a limitation not only of our work, but of all prior work in automatic literature review generation, going back to \citet{hoang-kan-2010-towards}. We suggest that future work can explore integrating citation list optimization with literature review generation, perhaps by generating iteratively candidate drafts and retrieving additional papers to cite.

\paragraph{Length limit of GPT-4.} Our experiments used gpt-4-0314, which has a maximum input token length of 8k; we did not have access to gpt-4-32k, which has four times the length limit. For nearly half of the related work sections, we had to iteratively generate subsections and concatenate the outputs, as described in Section \ref{sec:implementation_details}. Consequently, the coherence between subsections is significantly impacted, and they read like a concatenation of different related work sections. This problem will be mitigated as more LLMs have longer maximum input lengths.

\paragraph{Imperfect preprocessing pipeline.} We were unable to access the PDFs of some cited works due to several problems: 
  (1) the PDF parser is only able to parse research papers, but not books or websites; (2) the citation lists are automatically extracted from the parsed PDFs, and some papers may be missing; (3) we were unable to retrieve some cited papers using the Google seach API; and (4) we were unable to download some cited papers due to publisher pay-walls. The missing cited papers limit the performance of our system.
  
\paragraph{Inconsistent citation markers.} Since we use heuristics to extract citation markers from a JSON parsed from the target paper PDF, some of the author last names and publication years may not be accurate. In addition, we observe a few cases of inconsistency in citation styles (e.g. mixing ``Smith et al. (2023)'' and ``[1]'') across multiple passes of generation. In future work, we will leverage the LLM's code comprehension and generation ability to directly input the bibliography and output related work texts in \LaTeX format.

\paragraph{Quality of the intermediate outputs.} Since there are no gold features introduced in Section \ref{sec:citation_network_features} \& \ref{sec:citing_features}, nor do we have the resource of additional human evaluation for these intermediate output features from the LLM, we have to leave the study of intermediate feature quality and its influence to the final output for future work.

\paragraph{Post-processing layer.} In this preliminary study, we only limit our scope to the initial generation process without additional post-processing steps. We leave additional fact-checking and correction, and potential plagiarism avoidance for future work.

\paragraph{Proprietary LLM APIs.} Our prompts are designed based on OpenAI gpt-3.5-turbo-0301 and gpt-4-0314. As these models may be deprecated, the results may not be replicated in the future. Moreover, the prompts may have to be updated to adapt to newly released models. Nonetheless, we argue that the key input features and general prompt format we propose should be consistently useful across any LLM. We later test our prompts on other LLMs: gpt-3.5-turbo-0613 \& gpt-4-0613, as well as Anthropic Claude-v2 \footnote{\url{https://www.anthropic.com/index/claude-2}} output qualitatively similar texts, while Google text-bison-32k \footnote{\url{https://cloud.google.com/vertex-ai/docs/generative-ai/model-reference/text}} output texts with less satisfactory styles. On the other hand, LLaMA-2 70B Chat \cite{touvron2023llama} fails the task by generating related work sections irrelevant to the input.

\paragraph{Limited field of studies and generalizability.} As Appendix Table \ref{tab:field} shows, our evaluated papers are concentrated on the field of computer science, particularly natural language processing, and our prompts are also designed for computer science papers. We note that it is very challenging and expensive to use authors/experts to evaluate literature reviews and we do not have the resources to tune the LLM prompts and recruit experts in other disciplines. We do evaluate a geology paper because the expert is friends with an author, and we do not see any significant difference from the computer science evaluations. We leave other domains of target papers, and more dynamic prompt template 
for future work.

\section*{Ethics Statement}
As an early exploratory work, we use LLMs to automatically generate literature reviews. LLMs may produce inappropriate outputs, such as toxic or non-factual statements. LLMs may also plagiarize the cited papers; however, in our intended use case, generating a literature review summarizing a daily paper feed, this is less of a concern, since the review is shown only to feed owner for the purpose of assisting them in curating their reading list. 

Because our experiments are conducted using related work sections as evaluation targets, it is possible that unscrupulous individuals may use our system to ``cheat'' at writing related work sections for their own publications. We strongly advise against doing so, as this violates the requirement of a fully original piece of work for academic venues. Since there is not yet an established norm around the use of generative systems in writing scientific papers, there may be some risk of harm to the scientific community from careless use of such tools, and their use might be explicitly prohibited in some contexts. Therefore, future researchers, developers, and users must be extra careful about the potential regulations.

From a practical standpoint, the quality of literature reviews generated using our approach is still noticeably lower than human-written ones, especially in terms of organization and writing style. In addition, the human-provided main idea plan is required for higher-quality output, and the fully-automated setting performs very poorly, which should discourage the malicious use of our system. Our results clearly show that the human thinking process cannot be replaced by an automated system, and human readers are easily able to distinguish and criticize AI-generated content. 


\clearpage
\bibliography{anthology,custom}
\bibliographystyle{acl_natbib}
\clearpage
\appendix

\begin{table}[t]
\small
\begin{center}
\setlength{\tabcolsep}{5pt} 
    \begin{tabular}{p{0.9\linewidth} }
    \hline
    \textbf{Prompt} \\ \hline
    Title: \{\{title\}\} \\
    Abstract: \{\{abstract\}\} \\
    Introduction: \{\{introduction\}\} \\
    Conclusion: \{\{conclusion\}\} \\
    What are the objective, method, findings, contributions and keywords of the paper above? Answer in the format of \\ 
    Objective: XXX. \\ 
    Method: XXX. \\ 
    Findings: XXX. \\ 
    Contribution: XXX. \\ 
    Keywords: A; B; C. \\ \hline
    \textbf{Faceted Summary}\\ \hline
    Objective: \{\{objective\}\} \\ 
    Method: \{\{method\}\} \\ 
    Findings: \{\{findings\}\} \\ 
    Contribution: \{\{contribution\}\} \\ 
    Keywords: \{\{keywords\}\}  \\
   \hline
    \end{tabular}
    \vspace{-0.5em}
    \caption{Prompt and output format for generating faceted summary of a paper.} \label{tab:faceted_summary_prompt}
    \vspace{-1.5em}
\end{center}
\end{table}

\begin{table}[t]
\small
\begin{center}
\setlength{\tabcolsep}{5pt} 
    \begin{tabular}{p{0.9\linewidth} }
    \hline
    \textbf{Prompt} \\ \hline
    Faceted summary of the citing paper, \{\{title A\}\} by \{\{author A\}\} et al. \{\{year A\}\}:\\
    \{\{\emph{Faceted Summary A}\}\} \\ 
    Faceted summary of the cited paper, \{\{title B\}\} by \{\{author B\}\} et al. \{\{year B\}\}:\\
    \{\{\emph{Faceted Summary B}\}\} \\ 
    Citation contexts that \{\{author A\}\} et al. \{\{year A\}\} cites \{\{author B\}\} et al. \{\{year B\}\} (which is cited as \{\{citation marker of B in A\}\})): \\
    1. \{\{span \#1\}\} \\
    2. \{\{span \#2\}\} \\
    ...... \\
    Very briefly explain the relationship between \{\{author A\}\} et al. \{\{year A\}\} and \{\{title B\}\} by \{\{author B\}\} et al. \{\{year B\}\}. TLDR: \\ \hline
    \textbf{Relation Between Paper Pairs}\\ \hline
    \{\{author A\}\} et al. \{\{year A\}\} cites \{\{author B\}\} et al. \{\{year B\}\} ...... \\
    \hline
    \end{tabular}
    \vspace{-0.5em}
    \caption{Prompt and output format for generating the relationship between paper pairs. The ``citation marker of B in A'' is how paper A refers to paper B, e.g. ``B et al. (2023)'' or simply ``[1]''.} \label{tab:relationship_prompt}
    \vspace{-1.5em}
\end{center}
\end{table}

\begin{table}[t]
\small
\begin{center}
\setlength{\tabcolsep}{5pt} 
    \begin{tabular}{p{0.9\linewidth} }
    \hline
    \textbf{Prompt} \\ \hline
    How other papers cite \{\{author B\}\} et al. \{\{year B\}\}:\\
    \{\{\emph{Relation between A1 and B}\}\} \\ 
    Example citation fragments: \\
    1. \{\{span \#1 of A1 citing B\}\} \\
    2. \{\{span \#2 of A1 citing B\}\} \\
    ...... \\
    \{\{\emph{Relation between A2 and B}\}\} \\ 
    Example citation fragments: \\
    1. \{\{span \#1 of A2 citing B\}\} \\
    2. \{\{span \#2 of A2 citing B\}\} \\
    ...... \\
    ...... \\
    Very briefly answer what \{\{author B\}\} et al. \{\{year B\}\} is mostly known for, and the common citation intent. Hint: pay attention to how \{\{author B\}\} et al. \{\{year B\}\} is referred by the citing papers. Answer in the format of "\{\{author B\}\} et al. \{\{year B\}\} is known for XXX and it is cited for YYY". TLDR:  \\ \hline
    \textbf{Enriched Citation Usage}\\ \hline
    \{\{author B\}\} et al. \{\{year B\}\} is known for ... and it is cited for ... \\
    \hline
    \end{tabular}
    \vspace{-0.5em}
    \caption{Prompt and output format for generating enriched citation intent and usage of cited papers.} \label{tab:usage_prompt}
    \vspace{-1.5em}
\end{center}
\end{table}

\begin{table}[t]
\small
\begin{center}
\setlength{\tabcolsep}{5pt} 
    \begin{tabular}{p{0.9\linewidth} }
    \hline
    \textbf{Prompt} \\ \hline
    Our title: \{\{title\}\} \\
    Faceted summary of our paper: \\
     \{\{\emph{Faceted Summary}\}\} \\ 
    Write a short summary of the main idea of the following related work section paragraphs. Ignore citations. \\
    \{\{\emph{Human-written related work section}\}\} \\ \hline
    \textbf{Main idea of the target related work section}\\ \hline
    \{\{\emph{main ideas}\}\} \\
    \hline
    \end{tabular}
    \vspace{-0.5em}
    \caption{Prompt and output format for generating the main ideas of the target related work section.} \label{tab:main_idea_prompt}
    \vspace{-1.5em}
\end{center}
\end{table}

\begin{table}[t]
\small
\begin{center}
\setlength{\tabcolsep}{5pt} 
    \begin{tabular}{p{0.8\linewidth} }
    \hline
    \textbf{Prompt} \\ \hline
    We have finished writing the title, abstract, introduction and conclusion section of our NLP paper as follows: \\
    Title: \{\{title\}\} \\
    Abstract: \{\{abstract\}\} \\
    Introduction: \{\{introduction\}\} \\
    Conclusion: \{\{conclusion\}\} \\
    However, the related work section is still missing.
Write our related work section that concisely cites the following papers in a natural way using all of the main ideas as the main story.
Keep it short, e.g. 3 paragraphs at most. Make sure the related work section does not conflict with the sections already written.
You can freely reorder the cited papers to adapt to the main ideas.
Pay extra attention to <Usage> which indicates how each work is cited by other work.  \\ \hdashline
  Main idea of our related work section: \\ 
  \{\{\emph{main ideas}\}\} \\ \hdashline
    List of cited papers: \\
    1. \{\{title B1\}\} by \{\{author B1\}\} et al. \{\{year B1\}\} \\
    \{\{\emph{Faceted Summary or Abstract of B1}\}\} \\ 
    <Usage> \{\{Enriched citation usage of B1\}\} \\
    How other papers cite it: \\
    \{\{Relation between Ax and B1\}\} \\
    \{\{Relation between Ay and B1\}\} \\
    ... \\
    \hdashline
    Potentially useful sentences from this paper: 
    \{\{section \#1\}\} \{\{CTS \#1\}\} \\
    \{\{section \#2\}\} \{\{CTS \#2\}\} \\
    ... \\ \hdashline
    2. \{\{title B2\}\} by \{\{author B2\}\} et al. \{\{year B2\}\} \\
    ...... \\
    \hline
    \textbf{Output}\\ \hline
    \{\{\emph{related work section}\}\} \\
    \hline
    \end{tabular}
    \vspace{-0.5em}
    \caption{Prompt and output format for generating the full target related work section. Cited papers are given in chronological order.} \label{tab:full_related_work_generation_prompt}
    \vspace{-1.5em}
\end{center}
\end{table}

\begin{table}[t]
\small
\begin{center}
\setlength{\tabcolsep}{2.5pt} 
\renewcommand{\arraystretch}{1.0} 
    \begin{tabular}{lc}
    \hline
    \textbf{Field of Study} & \textbf{Count} \\ 
    \hline
    Natural Language Processing & 14 \\
    Machine Learning & 4 \\
    Speech and Audio Processing & 3 \\
    Computer Vision & 2 \\
    Programming Languages & 1 \\
    Robotics & 1 \\
    Computer Graphics & 1 \\
    Geoscience & 1 \\
    \hline
    \end{tabular}
    \vspace{-0.5em}
    \caption{Distribution of the field of study among the human-rated related work sections.} \label{tab:field}
    \vspace{-1.5em}
\end{center}
\end{table}

\begin{table}[t]
\small
\begin{center}
\setlength{\tabcolsep}{2.5pt} 
\renewcommand{\arraystretch}{1.0} 
    \begin{tabular}{lc}
    \hline
    \textbf{Year} & \textbf{Count} \\ 
    \hline
    2022 & 13 \\
    2023 & 7 \\
    2021 & 4 \\
    2019 & 1 \\
    2018 & 1 \\
    2016 & 1 \\
    \hline
    \end{tabular}
    \vspace{-0.5em}
    \caption{Distribution of the publication year among the human-rated related work sections.} \label{tab:year}
    \vspace{-1.5em}
\end{center}
\end{table}

\begin{table*}[t]
\small
\begin{center}
\setlength{\tabcolsep}{6pt} 
\renewcommand{\arraystretch}{1.0} 
    \begin{tabular}{ccccccccc}
    \hline
    \textbf{Metrics} & \textbf{A} & \textbf{B} & \textbf{C} & \textbf{D} & \textbf{E} & \textbf{F} & \textbf{G} & \textbf{H} \\ \hline
    \tablerow{Fluency}{$3.78$}{$4.11$}{$4.07$}{$4.11$}{$4.0$}{$4.19$}{$\mathbf{4.33}$}{$4.15$} \\
    \tablerow{}{$(1.23)$}{$(0.87)$}{$(0.81)$}{$(0.74)$}{$(0.9)$}{$(0.72)$}{$\mathbf{(0.72)}$}{$(0.59)$} \\ \hline
    \tablerow{Organization and Coherence}{$3.07$}{$3.30$}{$3.59$}{$3.59$}{$3.33$}{$3.37$}{$\mathbf{3.70}$}{$3.52$} \\
    \tablerow{}{$(1.15)$}{$(1.12)$}{$(0.91)$}{$(0.95)$}{$(0.90)$}{$(0.99)$}{$\mathbf{(0.90)}$}{$(0.96)$} \\ \hline
    \tablerow{Relevance (to target paper)}{$3.67$}{$3.78$}{$4.11$}{$4.00$}{$3.89$}{$4.00$}{$\mathbf{4.19}$}{$4.07$} \\
    \tablerow{}{$(1.09)$}{$(0.99)$}{$(0.83)$}{$(0.94)$}{$(1.03)$}{$(0.98)$}{$\mathbf{(0.67)}$}{$(0.81)$} \\ \hline
    \tablerow{Relevance (to cited papers)}{$3.93$}{$\mathbf{4.22}$}{$4.19$}{$4.19$}{$4.15$}{$4.04$}{$\mathbf{4.22}$}{$4.00$} \\
    \tablerow{}{$(1.05)$}{$\mathbf{(0.57)}$}{$(0.67)$}{$(0.77)$}{$(0.70)$}{$(0.74)$}{$\mathbf{(0.63)}$}{$(0.86)$} \\ \hline
    \tablerow{Factuality}{$3.89$}{$4.04$}{$3.93$}{$\mathbf{4.30}$}{$3.74$}{$3.74$}{$3.89$}{$3.93$} \\
    \tablerow{}{$(1.17)$}{$(0.96)$}{$(1.09)$}{$\mathbf{(0.76)}$}{$(1.20)$}{$(1.00)$}{$(1.10)$}{$(1.05)$} \\ \hline
    \tablerow{Usefulness/Informativeness}{$3.30$}{$3.74$}{$\mathbf{3.85}$}{$3.70$}{$3.59$}{$3.78$}{$3.52$}{$3.59$} \\
    \tablerow{}{$(1.01)$}{$(0.89)$}{$\mathbf{(0.89)}$}{$(1.08)$}{$(0.87)$}{$(0.68)$}{$(0.96)$}{$(0.95)$} \\ \hline
    \tablerow{Writing Style}{$3.07$}{$3.48$}{$3.70$}{$3.52$}{$3.30$}{$\mathbf{3.81}$}{$3.78$}{$3.44$} \\
    \tablerow{}{$(1.09)$}{$(0.88)$}{$(0.90)$}{$(0.83)$}{$(0.94)$}{$\mathbf{(0.90)}$}{$(0.74)$}{$(0.92)$} \\ \hline
    \tablerow{Overall}{$2.89$}{$3.33$}{$\mathbf{3.67}$}{$3.56$}{$3.15$}{$3.15$}{$3.48$}{$3.22$} \\
    \tablerow{}{$(0.96)$}{$(1.05)$}{$\mathbf{(0.82)}$}{$(1.03)$}{$(1.01)$}{$(0.93)$}{$(0.96)$}{$(0.96)$} \\ \hline
    \tablerow{\# of factual errors}{$0.78$}{$0.70$}{$0.81$}{$\mathbf{0.44}$}{$0.78$}{$0.89$}{$0.70$}{$0.74$} \\
    \tablerow{}{$(1.07)$}{$(0.94)$}{$(0.98)$}{$\mathbf{(0.92)}$}{$(1.07)$}{$(1.13)$}{$(0.81)$}{$(1.14)$} \\ \hline
    \end{tabular}
    \vspace{-0.5em}
    \caption{Average (and standard deviation) of human evaluation scores.} \label{tab:full_human_evaluation}
    \vspace{-1.5em}
\end{center}
\end{table*}

\begin{table}[t]
\small
\begin{center}
\setlength{\tabcolsep}{6pt} 
\renewcommand{\arraystretch}{1.0} 
    \begin{tabular}{ccccccccccc}
    \hline
    \textbf{Metrics} & \textbf{Average} & \textbf{Std} \\ \hline
    Fluency & 4.48 & 0.83 \\ \hline
    Organization and coherence & 4.15 & 0.80 \\ \hline
    Relevance-citing & 4.48 & 0.57 \\ \hline
    Relevance-cited & 4.52 & 0.50 \\ \hline
    Factuality & 4.44 & 0.62 \\ \hline
    \# of factual errors & 0.30 & 0.71 \\ \hline
    Usefulness/informativeness & 4.33 & 0.72 \\ \hline
    Writing style & 4.19 & 0.67 \\ \hline
    Overall & 4.41 & 0.62 \\ \hline    
    \end{tabular}
    \vspace{-0.5em}
    \caption{Average and standard deviation of the variation with the best human evaluation overall score.} \label{tab:best_human_evaluation}
    \vspace{-1.5em}
\end{center}
\end{table}

\begin{figure*}
     \centering
     \begin{subfigure}[b]{0.45\textwidth}
         \centering
         \includegraphics[width=\textwidth]{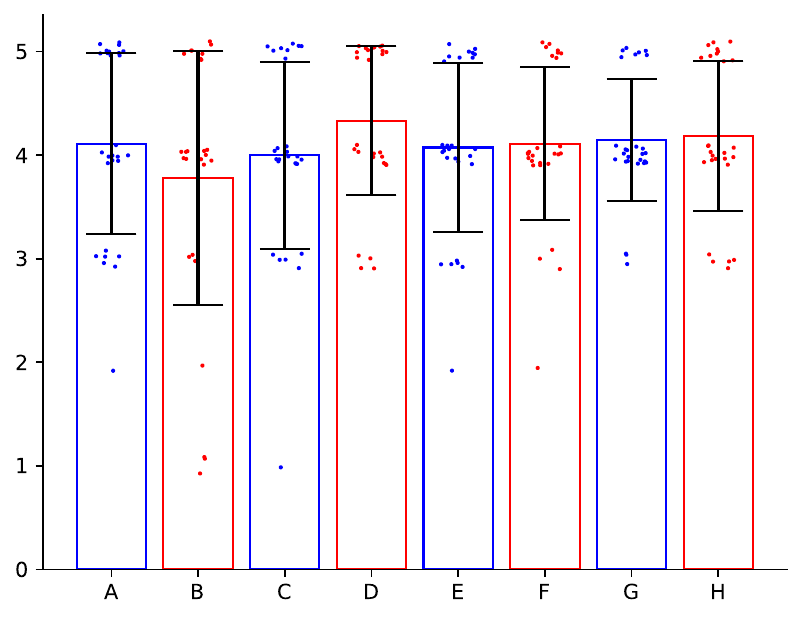}
         \vspace{-2.2em}
         \caption{Fluency}
         \label{fig:fluency}
     \end{subfigure}
     \quad
     \begin{subfigure}[b]{0.45\textwidth}
         \centering
         \includegraphics[width=\textwidth]{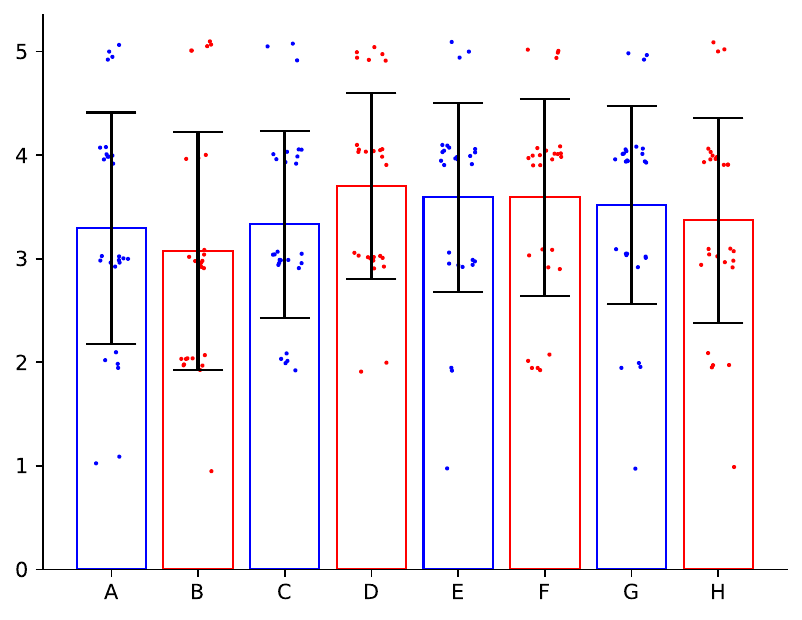}
         \vspace{-2.2em}
         \caption{Organization and coherence}
         \label{fig:coherence}
     \end{subfigure}

     \centering
     \begin{subfigure}[b]{0.45\textwidth}
         \centering
         \includegraphics[width=\textwidth]{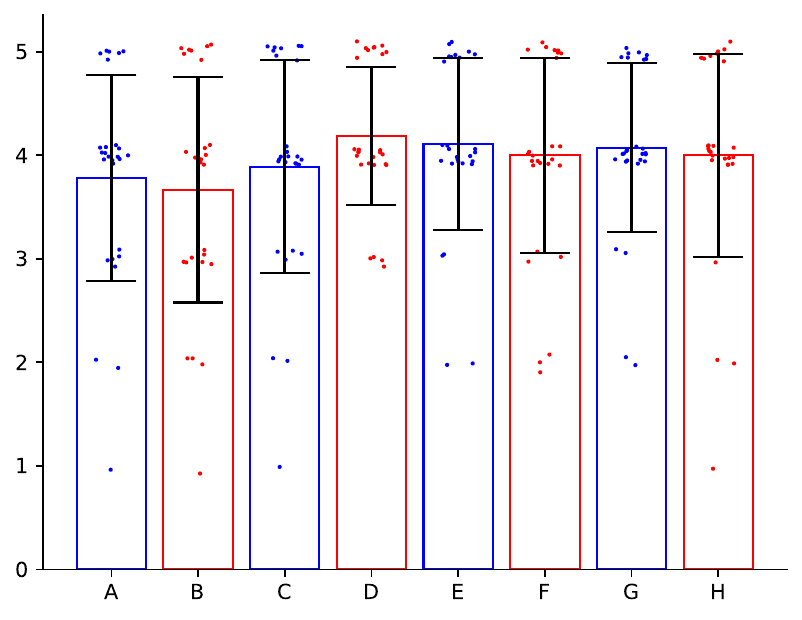}
         \vspace{-2.2em}
         \caption{Relevance to the target paper}
         \label{fig:relevance_citing}
     \end{subfigure}
     \quad
     \begin{subfigure}[b]{0.45\textwidth}
         \centering
         \includegraphics[width=\textwidth]{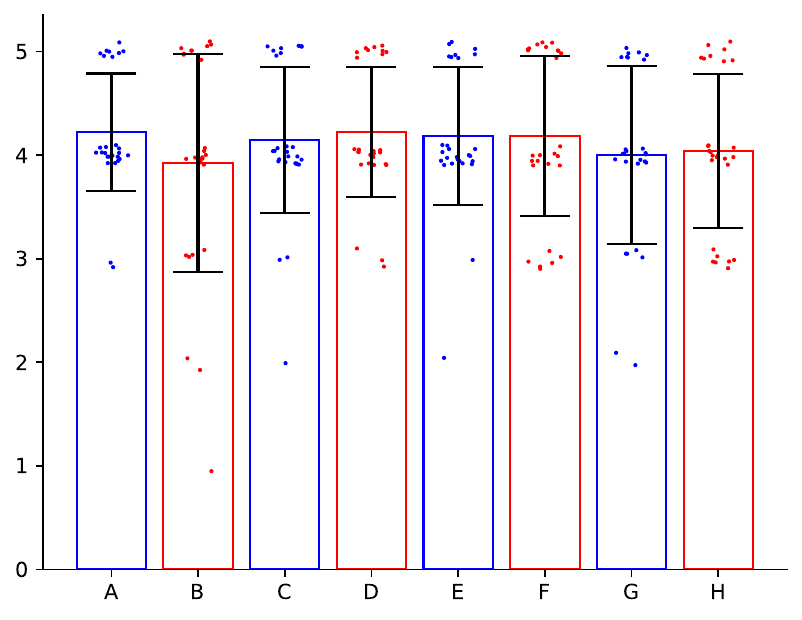}
         \vspace{-2.2em}
         \caption{Relevance to the cited paper}
         \label{fig:relevance_cited}
     \end{subfigure}

      \centering
     \begin{subfigure}[b]{0.45\textwidth}
         \centering
         \includegraphics[width=\textwidth]{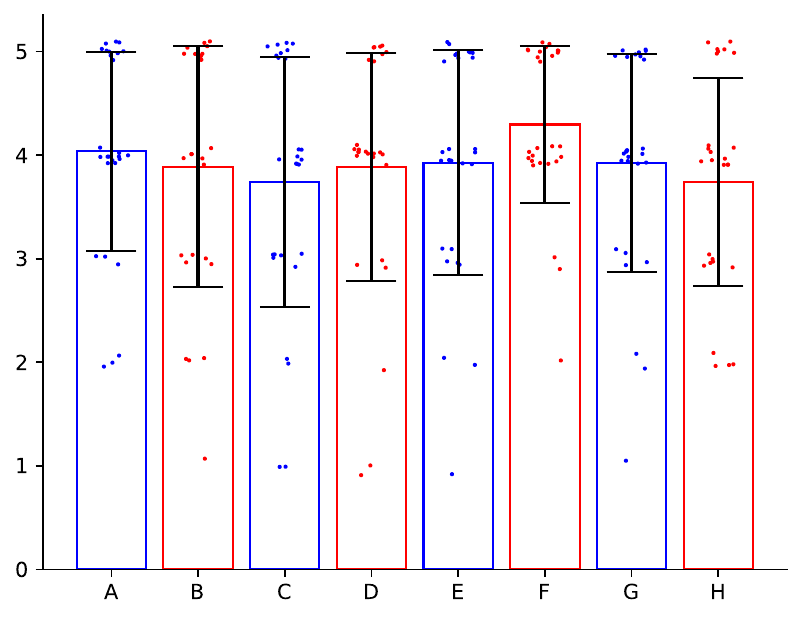}
         \vspace{-2.2em}
         \caption{Factuality}
         \label{fig:factuality}
     \end{subfigure}
     \quad
     \begin{subfigure}[b]{0.45\textwidth}
         \centering
         \includegraphics[width=\textwidth]{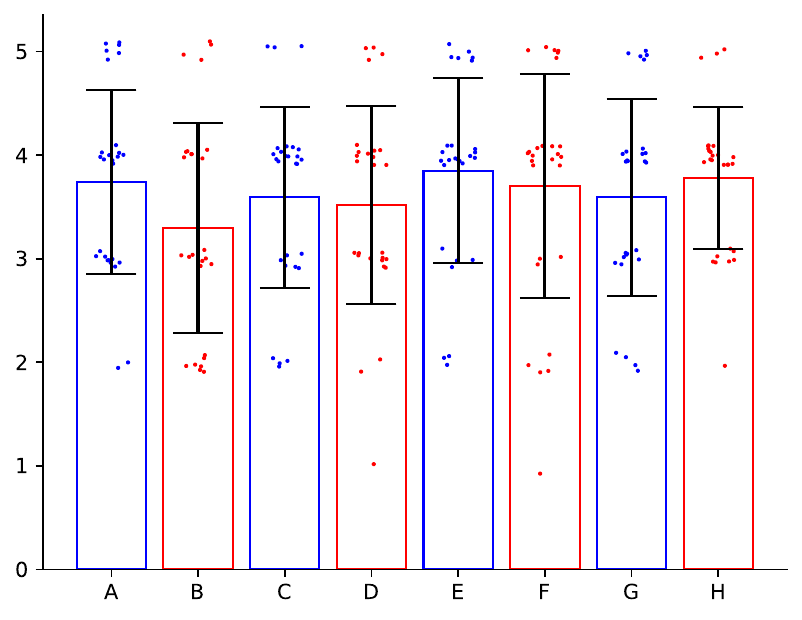}
         \vspace{-2.2em}
         \caption{Usefulness/informativeness}
         \label{fig:informativeness}
     \end{subfigure}

    \centering
     \begin{subfigure}[b]{0.45\textwidth}
         \centering
         \includegraphics[width=\textwidth]{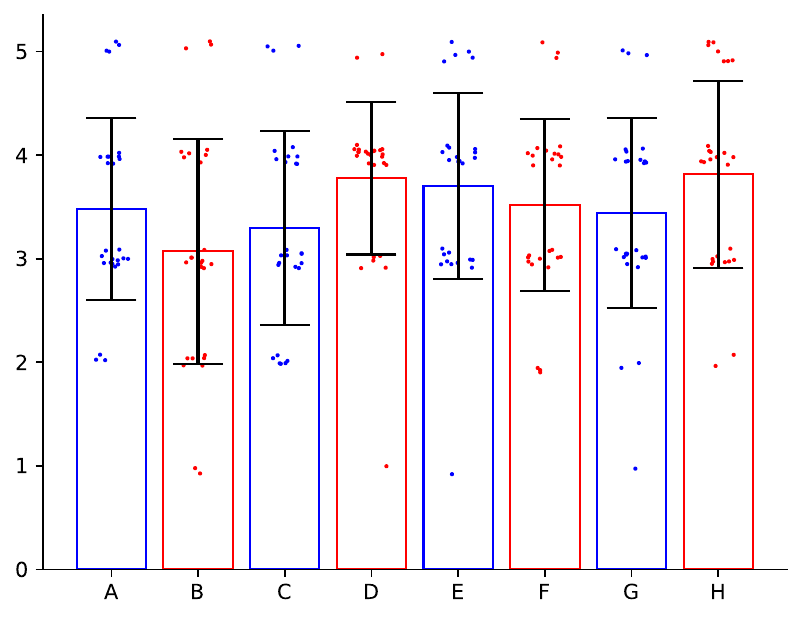}
         \vspace{-2.2em}
         \caption{Writing style}
         \label{fig:writing_style}
     \end{subfigure}
     \quad
     \begin{subfigure}[b]{0.45\textwidth}
         \centering
         \includegraphics[width=\textwidth]{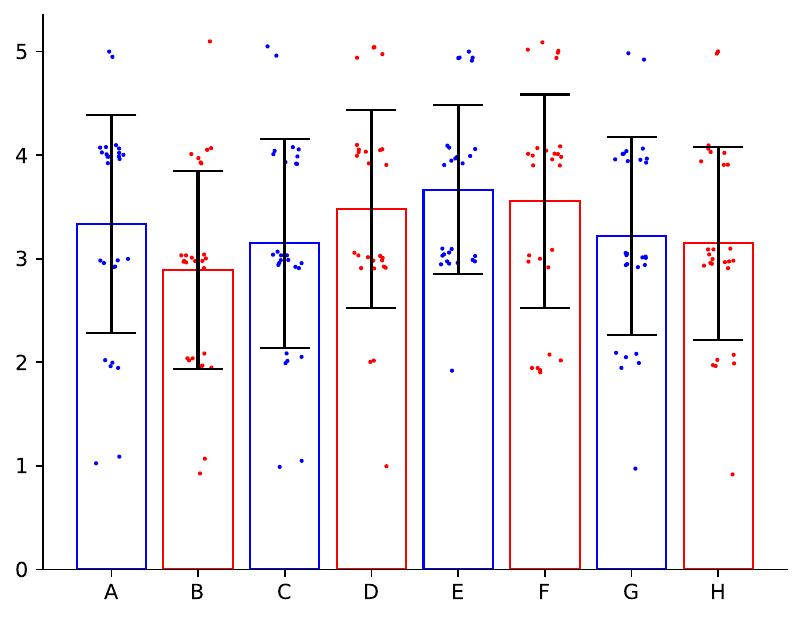}
         \vspace{-2.2em}
         \caption{Overall}
         \label{fig:overall}
     \end{subfigure}
  
     \vspace{0em}
        \caption{Bar chart showing the distribution of human evaluation scores.}
        \label{fig:human_evaluation}
\end{figure*}

\begin{figure}
\centering
  \includegraphics[width=0.45\textwidth]{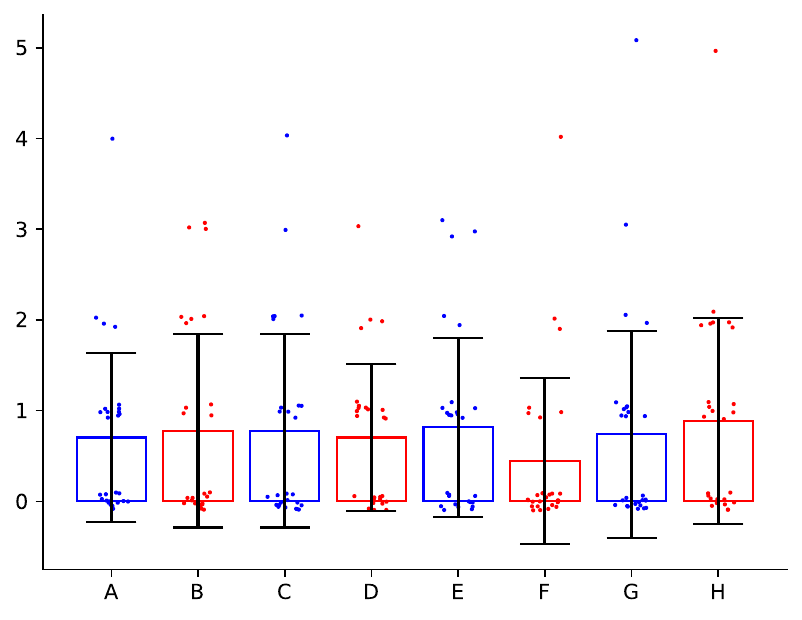}
  \vspace{-0.5em}
  \caption{Bar chart of the number of factual errors.} 
  \label{fig:factual_error}
  \vspace{-0.5em}
\end{figure}

\section{Extractiveness of Generations} \label{sec:component_analysis}

\begin{figure*}
     \centering
     \begin{subfigure}[b]{0.555\textwidth}
         \centering
         \includegraphics[width=\textwidth]{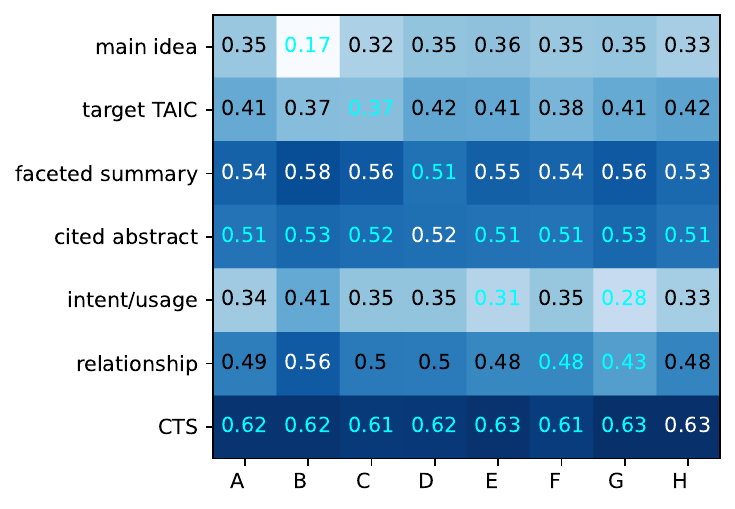}
         \vspace{-2.2em}
         \caption{Coverage}
         \label{fig:coverage}
     \end{subfigure}
     \quad
     \begin{subfigure}[b]{0.41\textwidth}
         \centering
         \includegraphics[width=\textwidth]{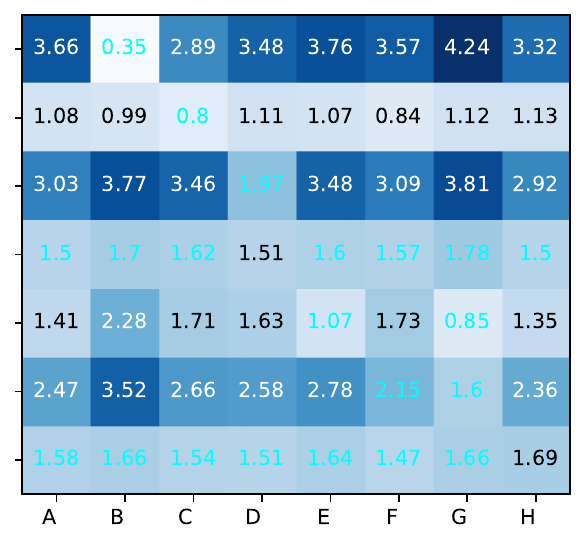}
         \vspace{-2.2em}
         \caption{Density}
         \label{fig:density}
     \end{subfigure}
  
     \vspace{-1.0em}
        \caption{Extractiveness of generated related work sections ($n=38$), measured by \textit{coverage} and \textit{density} against input features. Scores for features not included in the prompt for a variant are shown in \textcolor{cyan}{cyan}.}
        \label{fig:component_analysis}
        \vspace{-1.5em}
\end{figure*}

We use \citet{grusky-etal-2018-newsroom}'s \emph{coverage} and \emph{density} metrics to investigate how each of the features contributes to the generated related work sections. 

\paragraph{Coverage} \label{sec:coverage}
measures how much generated text is extracted from each of the input features. As Figure \ref{fig:coverage} shows, the features have very different coverage scores. Many, such as faceted summaries, cited paper abstracts, and CTS, have high coverage across all generation variants. Since the information among different features is likely to overlap, the coverage scores do not sum to 1 for each variant; for example, faceted summaries and cited paper abstracts contain highly similar information. 

Due to this overlap, certain features, such as cited paper abstracts and CTS, have high coverage scores even in variants where they are not part of the input prompt (cells in cyan), suggesting they contain information that is highly valued by the LLM. The coverage of each feature varies only slightly across prompt variants, likely due to the same information overlap reason. The only exception is variant $B$: the human-provided main ideas cannot be found in any other feature.

\paragraph{Density} \label{sec:density}
measures the length of extracted fragments. As the columns of Figure \ref{fig:density} show, scores vary significantly among features. Interestingly, despite faceted summaries and cited paper abstracts having similar coverage scores, the former have significantly higher density scores, indicating that the LLM prefers to directly copy from the faceted summaries. The main idea and relationship between paper pairs are also highly extractive. Note that, although this kind of extraction can be vulnerable to plagiarism, it is less of a concern for our approach because the highly extractive features are not taken directly from the cited papers, but from the feature-extraction LLM's summaries. 

Unlike coverage, density scores vary significantly among variants, since the LLM is not able to reconstruct an exact substring that is not present in its input. When features are ablated from a variant, their density scores are much lower (cells in cyan), and the scores of the remaining features are much higher. For example, in variant $G$, both the enriched citation usage and the relationship between paper pairs are ablated; to compensate, $G$ has very high density scores for the main idea and faceted summaries.

\section{Examples of Features and Generated Variants}
\label{sec:examples}

We use Section \ref{sec:related_work} of this paper as an example target related work section to demonstrate our key support features. For conciseness, we show only one example per feature, as showing all features across all cited papers would be extremely long.

We then generate the corresponding related work section variants A-H. Due to the maximum input length limit of our LLMs, we use the approach described in Section \ref{sec:full_related_work_generation} to generate the two paragraphs in each variant sequentially.

\subsection{Citation Network Features}

\paragraph{Faceted Summary of \citet{hoang-kan-2010-towards}.}

Objective: The objective of this paper is to introduce the problem of automated related work summarization and propose it as a challenge to the automatic summarization community. The paper aims to take initial steps towards solving this problem by dividing the task into general and specific summarization processes.

Method: The paper presents a prototype Related Work Summarization system, ReWoS, which takes in a set of keywords arranged in a hierarchical fashion that describes a target paper's topics, to drive the creation of an extractive summary using two different strategies for locating appropriate sentences for general topics as well as detailed ones. The paper also conducts a study of the argumentative patterns used in related work sections to describe the plausible summarization tactics for their creation.

Findings: The paper's initial results show an improvement over generic multi-document summarization baselines in a human evaluation.

Contribution: The paper introduces the novel problem of automated related work summarization and proposes it as a challenge to the automatic summarization community. The paper presents a prototype system, ReWoS, that takes initial steps towards solving this problem by dividing the task into general and specific summarization processes. The paper also conducts a study of the argumentative patterns used in related work sections to describe the plausible summarization tactics for their creation.

Keywords: automated summarization, related work summarization, multi-document summarization, citation prediction, keyphrase extraction.

\paragraph{Relationship between \citet{xing-etal-2020-automatic} and \citet{see-etal-2017-get}.} 

Xing et al. (2020) cite See et al. (2017) as a reference for the pointer-generator network architecture, which they adapt for their own task of automatic citation text generation in scholarly papers.

\paragraph{Enriched citation intent \& usage of \citet{xing-etal-2020-automatic}.} 

Xing et al. (2020) is known for proposing a task of automatic citation text generation in scholarly papers and developing a multi-source pointer-generator network with cross attention mechanism to address it. It is cited for related work and as a baseline for comparison in citation sentence generation tasks.

\subsection{Target Paper Features}

\paragraph{Main idea of Section \ref{sec:related_work}.} 

The paragraph discusses previous work on automatic related work generation, including Hoang and Kan's proposal of generating the related work section given a list of papers to cite, and later extractive approaches that use probabilistic models or BERT to select and concatenate sentences from cited papers. However, these approaches result in outputs lacking coherence and stylistic variation.

The paragraph discusses recent abstractive approaches that focus on citation text generation, which involves generating a single citation sentence or span given the cited paper(s) and assuming the rest of the related work section and other sections of the target paper are available. These approaches use various architectures and mostly rely on cited paper abstracts rather than full texts. However, there is no standard benchmark evaluation approach to compare methods from different prior works.

\subsection{CTS}

\paragraph{Extracted from \citet{hoang-kan-2010-towards}.}

[Structure of Related Work Section] Each article's data consists of the reference related work summary, the collection of the input research articles Inspired by the above observations, we propose a novel strategy for related work summarization with respect to a given topic tree.

\section{Example Generated Variants} \label{sec:example_related_work}
See Figures \ref{fig:variant_a}-\ref{fig:variant_h} for the reproduced Section \ref{sec:related_work} of this paper.  

\begin{figure*}[!htb]
\fbox{
\begin{minipage}{.95\textwidth}
In the last decade, several works have addressed the task of automatic related work generation. Hoang et al. (2010) introduced the problem of automated related work summarization and proposed a prototype system, ReWoS, which laid the foundation for subsequent works in the field. Hu et al. (2014) built upon this work by proposing a novel system for automatic related work generation using an optimization approach. Chen et al. (2019) further explored this area by proposing an approach for automatically generating a related work section in scientific papers through summarizing citations. Wang et al. (2019) introduced a new framework, ToC-RWG, for automatic related work generation by incorporating topic model and citation information. Deng et al. (2021) proposed a novel and effective system for generating related work sections based on sentence extraction and reordering using a BERT-based ensemble model. However, these approaches result in outputs lacking coherence and stylistic variation, which our work aims to address by proposing a workflow for generating a full related work section using an abstractive approach and prompting large language models with key support features. \\

Recent abstractive approaches have focused on citation text generation, which involves generating a single citation sentence or span given the cited paper(s) and assuming the rest of the related work section and other sections of the target paper are available. For instance, Abura'ed et al. (2020) proposed a method for generating related work sections in scientific documents using neural sequence learning, while Xing et al. (2020) introduced a task of automatic citation text generation in scholarly papers and developed a multi-source pointer-generator network with cross attention mechanism to address it. Ge et al. (2021) proposed a framework for citing sentence generation that considers both background knowledge and content information, and Luu et al. (2021) developed a dataset and model for explaining relationships between scientific documents using natural language text. Chen et al. (2021) proposed a relation-aware multi-document encoder for abstractive related work section generation. However, there is no standard benchmark evaluation approach to compare methods from different prior works.
\end{minipage}
}
\caption{Variant A (baseline).}
\label{fig:variant_a}
\end{figure*}

\begin{figure*}[!htb]
\fbox{
\begin{minipage}{.95\textwidth}
In the last decade, several works have addressed the task of automatic related work generation. Hoang et al. (2010) introduced the problem of automated related work summarization and proposed a prototype system, ReWoS, which laid the foundation for subsequent works in the field. Hu et al. (2014) built upon this work by proposing a novel system for automatic related work generation using an optimization approach. Chen et al. (2019) further explored the idea of generating related work sections by summarizing citations, while Wang et al. (2019) proposed a new framework, ToC-RWG, that incorporates topic model and citation information for related work generation. Deng et al. (2021) introduced a novel and effective system for generating related work sections based on sentence extraction and reordering using a BERT-based ensemble model. In contrast to these approaches, our work proposes a workflow for generating a full related work section by prompting large language models (LLMs) with key support features, aiming to produce a related work section that is comparable to a human first draft. \\

In recent years, several studies have focused on automatic related work generation. Abura'ed et al. (2020) proposed a method for generating related work sections in scientific documents using neural sequence learning. Xing et al. (2020) introduced a task of automatic citation text generation in scholarly papers and developed a multi-source pointer-generator network with cross attention mechanism to address it. Ge et al. (2021) proposed a framework for citing sentence generation that considers both background knowledge and content information. Luu et al. (2021) developed a dataset and model for explaining relationships between scientific documents using natural language text. Chen et al. (2021) proposed a relation-aware multi-document encoder for abstractive related work section generation. Li et al. (2022) developed a dataset for citation-oriented related work annotation and proposed a framework for automatic related work generation. These works have employed various techniques, such as Transformer models (Vaswani et al., 2017), pointer-generator networks (See et al., 2017), and large language models (Radford et al., 2019; Beltagy et al., 2020). Our work builds upon these studies by proposing a workflow for generating a related work section that is comparable to a human draft by extracting and integrating key support features into informative prompts for an LLM.
\end{minipage}
}
\caption{Variant B (main idea ablated).}
\label{fig:variant_b}
\end{figure*}

\begin{figure*}[!htb]
\fbox{
\begin{minipage}{.95\textwidth}
The problem of automated related work summarization was first introduced by Hoang et al. (2010), who proposed a prototype system, ReWoS, for generating related work sections. Subsequent works have built upon this foundation, such as Hu et al. (2014), who proposed an optimization approach for automatic related work generation using a PLSA model and regression models. Chen et al. (2019) introduced an approach for generating related work sections by summarizing citations, while Wang et al. (2019) proposed the ToC-RWG framework, which incorporates topic models and citation information. More recently, Deng et al. (2021) developed a BERT-based ensemble model for sentence extraction and reordering in related work sections. However, these extractive approaches often result in outputs lacking coherence and stylistic variation, motivating our work on generating a full related work section by prompting large language models and identifying key support features. \\

Recent abstractive approaches to citation text generation have focused on generating single citation sentences or spans, given the cited paper(s) and assuming the availability of the rest of the related work section and other sections of the target paper. Abura'ed et al. (2020) proposed a method for generating related work sections in scientific documents using neural sequence learning, while Xing et al. (2020) introduced a task of automatic citation text generation in scholarly papers and developed a multi-source pointer-generator network with cross attention mechanism to address it. Ge et al. (2021) proposed a framework for citing sentence generation that considers both background knowledge and content information, and Luu et al. (2021) developed a dataset and model for explaining relationships between scientific documents using natural language text. Chen et al. (2021) proposed a relation-aware multi-document encoder for abstractive related work section generation, and Li et al. (2022) developed a dataset for citation-oriented related work annotation and proposed a framework for automatic related work generation. However, there is no standard benchmark evaluation approach to compare methods from different prior works, and most of these approaches rely on cited paper abstracts rather than full texts.
\end{minipage}
}
\caption{Variant C (target TAIC ablated).}
\label{fig:variant_c}
\end{figure*}

\begin{figure*}[!htb]
\fbox{
\begin{minipage}{.95\textwidth}
In the last decade, several works have addressed the task of automatic related work generation. Hoang et al. (2010) introduced the problem of automated related work summarization and proposed a prototype system for it. Subsequent works, such as Hu et al. (2014), proposed a system for automatic related work generation using an optimization approach, while Chen et al. (2019) focused on generating related work sections through summarizing citations. Wang et al. (2019) proposed a framework for automatic related work generation by incorporating topic model and citation information. However, these extractive approaches often result in outputs lacking coherence and stylistic variation. Deng et al. (2021) addressed this issue by proposing a novel system for generating related work sections based on sentence extraction and reordering using a BERT-based ensemble model. In contrast, our work explores generating a full related work section using an abstractive approach by prompting large language models with key support features, aiming to produce a related work section that is comparable to a human first draft. \\

Recent abstractive approaches have focused on citation text generation, which involves generating a single citation sentence or span given the cited paper(s) and assuming the rest of the related work section and other sections of the target paper are available. Xing et al. (2020) proposed a task of automatic citation text generation in scholarly papers and developed a multi-source pointer-generator network with cross attention mechanism to address it. Ge et al. (2021) proposed a framework for citing sentence generation that considers both background knowledge and content information. Luu et al. (2021) developed a dataset and model for explaining relationships between scientific documents using natural language text, using citing sentences as a proxy for the task. Abura'ed et al. (2020) proposed a method for generating related work sections in scientific documents using neural sequence learning. Chen et al. (2021) proposed a relation-aware multi-document encoder for abstractive related work section generation. These approaches mostly rely on cited paper abstracts rather than full texts and use various architectures, such as Transformer models (Vaswani et al., 2017) and Longformer (Beltagy et al., 2020). However, there is no standard benchmark evaluation approach to compare methods from different prior works, as highlighted by Li et al. (2022), who conducted a meta-study on automatic related work generation.
\end{minipage}
}
\caption{Variant D (cited paper abstracts instead of faceted summaries).}
\label{fig:variant_d}
\end{figure*}

\begin{figure*}[!htb]
\fbox{
\begin{minipage}{.95\textwidth}
In the last decade, several works have addressed the task of automatic related work generation. Hoang et al. (2010) introduced the problem and proposed a prototype system, ReWoS, for automated related work summarization. Later, extractive approaches emerged, such as Hu et al. (2014), who proposed an optimization approach for automatic related work generation using a PLSA model, and Chen et al. (2019), who focused on summarizing citations for related work generation. Wang et al. (2019) proposed the ToC-RWG framework, which combined topic modeling and citation information for related work generation. Deng et al. (2021) introduced a BERT-based ensemble model for sentence extraction and reordering to generate related work sections. However, these extractive approaches often result in outputs lacking coherence and stylistic variation, motivating our work on a fully abstractive related work generation system using large language models. \\

Recent abstractive approaches have focused on citation text generation, which involves generating a single citation sentence or span given the cited paper(s) and assuming the rest of the related work section and other sections of the target paper are available. Xing et al. (2020) proposed a new task of automatic citation text generation in scholarly papers and developed a multi-source pointer-generator network with cross attention mechanism to address it. Ge et al. (2021) proposed a framework for citing sentence generation that considers both background knowledge and content information, while Abura'ed et al. (2020) designed and evaluated an abstractive related work section generation system. Luu et al. (2021) focused on explaining relationships between scientific documents using natural language text, using citing sentences as a proxy for the task. Chen et al. (2021) addressed the abstractive related work generation task and proposed a relation-aware multi-document encoder. Li et al. (2022) developed a dataset for citation-oriented related work annotation and proposed a framework for automatic related work generation. However, there is no standard benchmark evaluation approach to compare methods from different prior works.
\end{minipage}
}
\caption{Variant E (citation intent/usage ablated).}
\label{fig:variant_e}
\end{figure*}

\begin{figure*}[!htb]
\fbox{
\begin{minipage}{.95\textwidth}
The field of automatic related work generation has evolved significantly since Hoang et al. (2010) introduced the problem and proposed a prototype system, ReWoS, for automated related work summarization. Subsequent works have explored various approaches to tackle this challenge. Hu et al. (2014) proposed an optimization-based system, ARWG, which leverages a PLSA model and regression models to generate related work sections. Chen et al. (2019) introduced an approach that generates related work sections by summarizing citations, while Wang et al. (2019) presented a framework, ToC-RWG, that incorporates topic models and citation information for related work generation. More recently, Deng et al. (2021) proposed SERGE, a BERT-based ensemble model for sentence extraction and reordering to generate descriptive related work sections. Despite these advancements, existing extractive approaches often result in outputs lacking coherence and stylistic variation, motivating the need for more advanced abstractive methods, such as those based on BERT (Devlin et al., 2019), to further improve the quality of automatically generated related work sections. \\

Recent abstractive approaches have focused on citation text generation, which involves generating a single citation sentence or span given the cited paper(s) and assuming the rest of the related work section and other sections of the target paper are available. These approaches use various architectures and mostly rely on cited paper abstracts rather than full texts. For instance, Abura'ed et al. (2020) proposed a method for generating related work sections using neural sequence learning, while Xing et al. (2020) introduced a task of automatic citation text generation and developed a multi-source pointer-generator network with cross attention mechanism. Ge et al. (2021) proposed a framework for citing sentence generation that considers both background knowledge and content information. Luu et al. (2021) developed a dataset and model for explaining relationships between scientific documents using natural language text. Chen et al. (2021) proposed a relation-aware multi-document encoder for abstractive related work section generation. However, there is no standard benchmark evaluation approach to compare methods from different prior works. Some studies, such as Li et al. (2022), have conducted meta-studies on automatic related work generation, comparing existing literature and identifying potential areas for future research.
\end{minipage}
}
\caption{Variant F (relationship between papers ablated).}
\label{fig:variant_f}
\end{figure*}

\begin{figure*}[!htb]
\fbox{
\begin{minipage}{.95\textwidth}
In the last decade, several works have addressed the task of automatic related work generation. Hoang et al. (2010) introduced the problem of automated related work summarization and proposed a prototype system, ReWoS, which generates extractive summaries using keywords arranged in a hierarchical fashion. Hu et al. (2014) proposed an optimization approach, ARWG, which uses a PLSA model to split the sentence set of given papers into different topic-biased parts and employs regression models to learn the importance of sentences. Chen et al. (2019) presented an approach for generating related work sections by summarizing citations, while Wang et al. (2019) proposed a framework, ToC-RWG, that incorporates topic models and citation information for related work generation. More recently, Deng et al. (2021) developed SERGE, a BERT-based ensemble model for sentence extraction and reordering to generate descriptive related work sections. However, these extractive approaches often result in outputs lacking coherence and stylistic variation, motivating the need for more advanced abstractive methods. \\

Recent abstractive approaches have focused on citation text generation, which involves generating a single citation sentence or span given the cited paper(s) and assuming the rest of the related work section and other sections of the target paper are available. These approaches use various architectures and mostly rely on cited paper abstracts rather than full texts (Abura'ed et al., 2020; Xing et al., 2020; Ge et al., 2021; Luu et al., 2021; Chen et al., 2021). For instance, Abura'ed et al. (2020) designed and evaluated an abstractive related work section generation system, while Xing et al. (2020) proposed a multi-source pointer-generator network with a cross-attention mechanism for citation text generation. Ge et al. (2021) introduced a framework called BACO that considers both background knowledge and content information, and Luu et al. (2021) used citing sentences as a proxy for explaining relationships between scientific documents. Chen et al. (2021) proposed a Relation-aware Related work Generator (RRG) for abstractive related work section generation from multiple scientific papers. Li et al. (2022) developed a dataset for citation-oriented related work annotation and proposed a framework for automatic related work generation. However, there is no standard benchmark evaluation approach to compare methods from different prior works.
\end{minipage}
}
\caption{Variant G (citation intent/usage and relationship between papers ablated).}
\label{fig:variant_g}
\end{figure*}

\begin{figure*}[!htb]
\fbox{
\begin{minipage}{.95\textwidth}
The task of automatic related work generation has been explored in various ways over the past decade. Hoang et al. (2010) introduced the problem and proposed a prototype system, ReWoS, which laid the foundation for subsequent works. Hu et al. (2014) built upon this by proposing an optimization approach for related work generation, while Chen et al. (2019) focused on summarizing citations to generate related work sections. Wang et al. (2019) proposed the ToC-RWG framework, which incorporated topic models and citation information for related work generation. Deng et al. (2021) developed a BERT-based ensemble model for sentence extraction and reordering to generate related work sections. However, these approaches often result in outputs lacking coherence and stylistic variation, motivating our work on generating a full related work section using an abstractive approach with large language models. \\

In recent years, several abstractive approaches have focused on citation text generation, which involves generating a single citation sentence or span given the cited paper(s) and assuming the rest of the related work section and other sections of the target paper are available. For instance, Abura'ed et al. (2020) proposed a method for generating related work sections in scientific documents using neural sequence learning, while Xing et al. (2020) introduced a task of automatic citation text generation in scholarly papers and developed a multi-source pointer-generator network with cross attention mechanism to address it. Ge et al. (2021) proposed a framework for citing sentence generation that considers both background knowledge and content information, and Luu et al. (2021) developed a dataset and model for explaining relationships between scientific documents using natural language text. Chen et al. (2021) addressed the abstractive related work generation task and proposed a relation-aware multi-document encoder. However, there is no standard benchmark evaluation approach to compare methods from different prior works, and most of these approaches rely on cited paper abstracts rather than full texts.
\end{minipage}
}
\caption{Variant H (baseline with CTS enhancement).}
\label{fig:variant_h}
\end{figure*}

\end{document}